\documentclass[journal]{IEEEtran}
\usepackage{amsmath,amsfonts}
\usepackage{algorithmic}
\usepackage{algorithm}
\usepackage{array}
\usepackage[caption=false,font=normalsize,labelfont=sf,textfont=sf]{subfig}
\usepackage{textcomp}
\usepackage{stfloats}
\usepackage{url}
\usepackage{verbatim}
\usepackage{graphicx}
\usepackage{cite}
\usepackage{xcolor}
\usepackage{multirow}
\usepackage{multicol}
\usepackage{pifont}
\begin{document}

\title{A Physical Model-Guided Framework for Underwater Image Enhancement and Depth Estimation}
\author{Dazhao Du, Lingyu Si, Fanjiang Xu, Jianwei Niu,~\IEEEmembership{Senior Member,~IEEE}, and Fuchun Sun,~\IEEEmembership{Fellow,~IEEE}
\thanks{This work was supported by the Strategic Priority Research Program of the Chinese Academy of Sciences, Grant No. XDA 0370604. 
\emph{(Corresponding author: Fanjiang Xu)}
}
\thanks{Dazhao Du and Jianwei Niu are with Hangzhou Innovation Institute, Beihang University, Hangzhou, Zhejiang 310051, China (e-mail: dudazhao16@gmail.com; niujianwei@buaa.edu.cn); Lingyu Si and Fanjiang Xu are with the Institute of Software, Chinese Academy of Sciences, Beijing 100190, China (e-mail: lingyu@iscas.ac.cn; fanjiang@iscas.ac.cn); Fuchun Sun is with the Department of Computer Science and Technology, Tsinghua University, Beijing 100084, China (e-mail: fcsun@mail.tsinghua.edu.cn).}

}

\markboth{Journal of \LaTeX\ Class Files,~Vol.~14, No.~8, August~2021}%
{Shell \MakeLowercase{\textit{et al.}}: A Sample Article Using IEEEtran.cls for IEEE Journals}

\IEEEpubid{0000--0000/00\$00.00~\copyright~2021 IEEE}

\maketitle

\begin{abstract}
Due to the selective absorption and scattering of light by diverse aquatic media, underwater images usually suffer from various visual degradations. Existing underwater image enhancement (UIE) approaches that combine underwater physical imaging models with neural networks often fail to accurately estimate imaging model parameters such as scene depth and veiling light, resulting in poor performance in certain scenarios. To address this issue, we propose a physical model-guided framework for jointly training a Deep Degradation Model (DDM) with any advanced UIE model. DDM includes three well-designed sub-networks to accurately estimate various imaging parameters: a veiling light estimation sub-network, a factors estimation sub-network, and a depth estimation sub-network. Based on the estimated parameters and the underwater physical imaging model, we impose physical constraints on the enhancement process by modeling the relationship between underwater images and desired clean images, i.e., outputs of the UIE model. Moreover, while our framework is compatible with any UIE model, we design a simple yet effective fully convolutional UIE model, termed UIEConv. UIEConv utilizes both global and local features for image enhancement through a dual-branch structure. UIEConv trained within our framework achieves remarkable enhancement results across diverse underwater scenes. Furthermore, as a byproduct of UIE, the trained depth estimation sub-network enables accurate underwater scene depth estimation. Extensive experiments conducted in various real underwater imaging scenarios, including deep-sea environments with artificial light sources, validate the effectiveness of our framework and the UIEConv model.

\end{abstract}


\section{Introduction}


Underwater images often suffer from color distortion, low contrast, and blurriness caused by light absorption and scattering in water. These issues are further exacerbated by varying water conditions, such as turbidity and depth, which result in uneven lighting and haziness. By improving the visibility and clarity of underwater images, underwater image enhancement (UIE) facilitates the analysis, monitoring, and exploration of underwater scenes.

Early UIE methods simulate the degradation process by an underwater physical imaging model and estimate its parameters to invert clear images~\cite{carlevaris2010initial,drews2013transmission,galdran2015automatic,peng2015single}. With the introduction of paired datasets containing underwater images and reference images~\cite{uieb,uwcnn,utrans}, data-driven approaches have gradually gained attention. Numerous UIE models have been proposed to improve performance and efficiency~\cite{ugan,funiegan,ucolor,uiec,fivenet}. However, data-driven methods often suffer from poor generalization and interpretability. To combine the advantages of physical model-based methods and the powerful representational capacity of neural networks, some approaches~\cite{kar2021zero,yan2023hybrur} employ neural networks to estimate the parameters of the physical imaging model and then invert the degradation process based on the estimated parameters to obtain enhanced images. We argue that previous methods face two primary issues: (1) \textit{inaccurate imaging parameter estimation}; and (2) \textit{fundamental mismatches between idealized physical imaging models and real-world underwater conditions, making degradation inversion unreliable even with accurate estimation}. 

\IEEEpubidadjcol 

\begin{table*}[htbp]
\centering
\caption{Comparison of neural network-based parameter estimation for underwater imaging models. Existing methods either neglect certain parameters (marked N/A) or make simplifying assumptions, while our approach achieves more accurate estimation of all key parameters: scene depth, veiling light, and attenuation/scattering coefficients.}
\label{tab:method_comparison}
\begin{tabular}{lcccc}
\hline
Method  & Scene Depth & Veiling Light & Atten./Scatt. Coefficients & Coefficients Estimation \\
\hline
USUIR~\cite{fu2022unsupervised} & N/A & Uniform color image & Assumed equal & Transmission map \\
PUGAN~\cite{cong2023pugan} & Ground truth supervision & N/A & Assumed equal & Single-network \\
HybrUR~\cite{yan2023hybrur} & CycleGAN loss + rescaled to predefined range & Uniform color image & Assumed unequal & Dual-network \\
USeReDINet~\cite{varghese2023self} & Multi-view synthesis constraints & Uniform color image & Assumed unequal & Dual-network \\
\hline
Proposed Method & Physical constraint loss + dynamic rescaling & Non-uniform color image & Assumed unequal & Dual-network \\
\hline
\end{tabular}
\end{table*}

To address the first issue, we design a Deep Degradation Model (DDM), which includes three well-designed sub-networks to estimate scene depth, veiling light, attenuation coefficient, and scattering coefficient. We briefly outline the limitations of previous methods in estimating parameters and our solutions. (1) \textbf{Scene Depth:} To obtain absolute depth, relative depth is typically scaled based on manually set maximum and minimum depths. For example, in HybrUR~\cite{yan2023hybrur}, the minimum and maximum depth in each image are empirically set to 0.1m and 6m, respectively. These predefined depth ranges are unreasonable. Therefore, we propose a depth estimation sub-network to output accurate relative depth maps and a factors estimation sub-network to adaptively output scales and offsets for scaling relative depth values for each image. (2) \textbf{Veiling Light:} Veiling light, considered as the value of backscatter at infinity, is usually represented by a uniform color image where all pixel values are identical across spatial locations~\cite{fu2022unsupervised,varghese2023self,yan2023hybrur}. This assumption holds in shallow-water environments where light primarily originates from natural sources but fails in deep-sea scenes with artificial light sources, where underwater images often suffer from uneven lighting~\cite{marques2020l2uwe,hou2023non}. To adapt to various complex environments, we propose a veiling light estimation sub-network that outputs a local background light map where pixel values vary across spatial locations instead of a uniform color image. (3) \textbf{Attenuation and Scattering Coefficients}: When estimating scattering and attenuation coefficients, some methods directly assume they are identical~\cite{cong2023pugan}. Even if some methods estimate the transmission map instead of these two coefficients~\cite{kar2021zero,fu2022unsupervised}, they still rely on the assumption that both coefficients are identical. Recent work~\cite{akkaynak2018revised} has experimentally shown that they are related but not identical. Therefore, our factors estimation sub-network employs two separate heads to distinguish between them, providing a more accurate modeling approach. Table~\ref{tab:method_comparison} delineates the distinctions between our method and existing approaches in modeling these imaging parameters.

\begin{figure*}[t]
\centering
\includegraphics[width=6.8in]{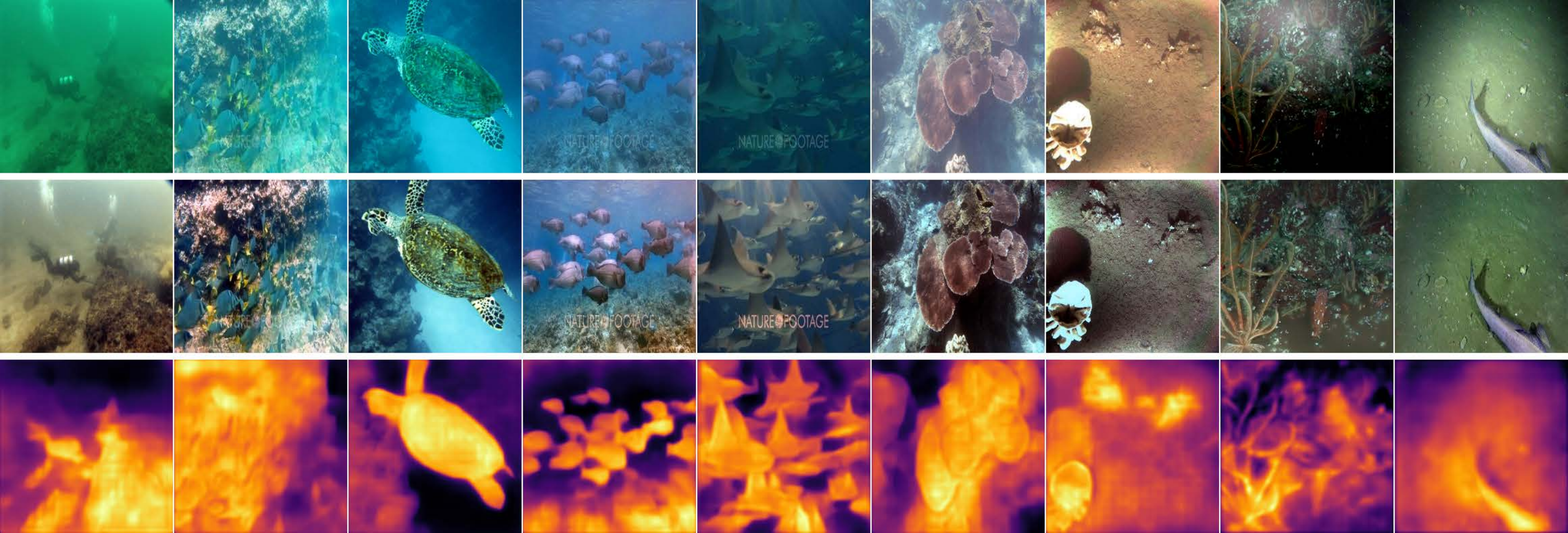}
\caption{For the original underwater images with various degradation effects, including bluish, greenish, yellowish hues, turbidity, blur, haziness, and uneven lighting (first row), we visualize the enhancement results (second row) and depth estimation results (third row) obtained by our method. Our method demonstrates impressive performance across various complex underwater environments.}
\label{fig:examples}
\end{figure*}

To address the second issue, we do not directly use the estimated imaging parameters to invert clean images, as done by previous methods. Instead, we use the physical imaging model to degrade the enhanced image produced by the UIE model. We enforce that the degraded enhanced image closely resembles the original underwater image, thereby imposing physical constraints on the enhancement process of the UIE model, as shown in Fig.~\ref{fig:framework}. In our training framework, physical constraints manifest as an additional regularization term that assists in the training of the UIE model. In this way, the UIE model with strong fitting capabilities can further be guided and supplemented by domain knowledge of the imaging process. Furthermore, considering that the scene depth of the original and enhanced underwater images should remain consistent, we introduce a novel depth consistency loss to further assist in the training process.

While our framework is compatible with arbitrary advanced UIE models, we specifically propose a novel UIE model, UIEConv, that offers superior simplicity and effectiveness compared to existing models. Effective image restoration hinges on capturing both local and global features. Global features are critical for correcting overarching aspects such as color and lighting, whereas local features are essential for restoring intricate high-frequency details. Recognizing the inherent limitations of convolutional receptive fields, some models employ frequency domain operations~\cite{fivenet} or Transformers~\cite{utrans} to extract global information. Our innovation lies in UIEConv, a simple yet powerful fully convolutional UIE model. It uniquely features distinct global and local branches. The global branch, designed with a U-Net-like architecture, progressively downsamples to capture high-level features and subsequently upsamples to reconstruct the image. This design inherently provides a large receptive field, crucial for modeling long-range dependencies. Complementing this, the local branch, with its more constrained receptive field, utilizes several convolutions to preserve image resolution throughout. The final enhanced result is obtained by combining the outputs of the two branches.

Our main contributions can be summarized as follows:
\begin{itemize}
    \item We propose a physical model-guided training framework that jointly performs image enhancement and depth estimation, with the two tasks complementing and enhancing each other. Additionally, any advanced UIE model can be trained within this framework to achieve further performance improvements.
    \item Within our framework, we meticulously design various sub-networks to accurately estimate crucial parameters for the underwater physical imaging model, including veiling light, scene depth, attenuation coefficient, and scattering coefficient.
    \item We introduce a simple yet effective UIE model, termed UIEConv. It employs a dual-branch structure to fully exploit both global and local information in the image. Without any bells and whistles, UIEConv outperforms other state-of-the-art UIE models.
    \item We test our method in various underwater scenarios, including different water bodies and deep-sea scenes with limited lighting. As shown in Fig.~\ref{fig:examples}, our approach consistently achieves impressive enhancement and depth estimation results. Extensive experiments validate the effectiveness of our approach.
\end{itemize}



\section{Related Work}

\subsection{Underwater Physical Imaging Model}

\begin{figure}[!t]
\centering
\includegraphics[width=3.1in]{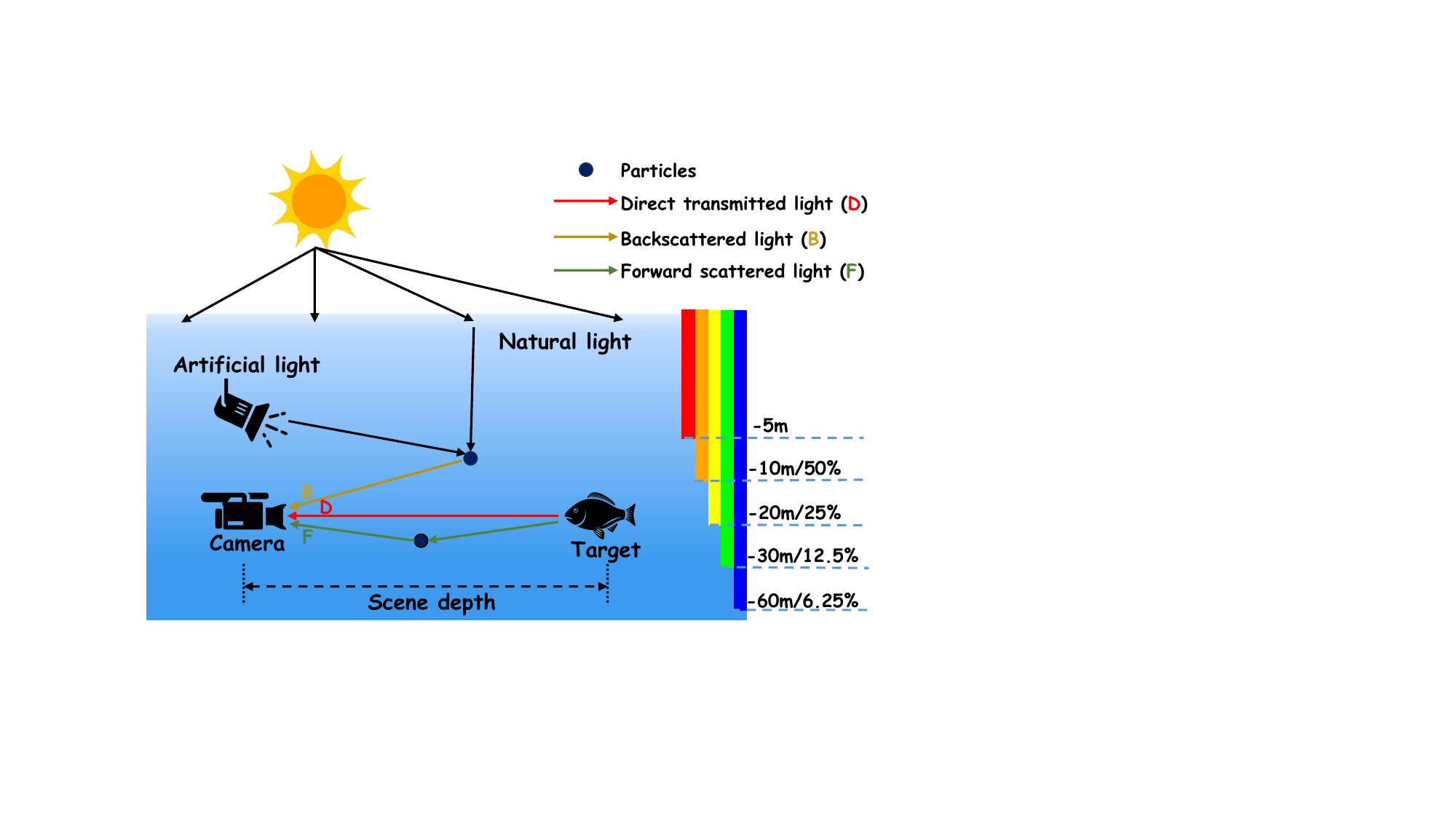}
\caption{Schematic diagram of underwater imaging process.}
\vspace{-0.1cm}
\label{fig:physicalmodel}
\end{figure}

According to~\cite{akkaynak2017space}, as shown in Fig.~\ref{fig:physicalmodel}, the light entering the camera mainly consists of three components: backscattered light $B$, direct transmitted light $D$, and forward scattered light $F$. Given that $F\ll D$, forward scattered light does not significantly contribute to the degradation of an image. Therefore, the underwater physical imaging model can be represented as the sum of direct transmitted light $D$ and backscattered light $B$:
\begin{subequations}
\label{eq:imagemodel}
\begin{align}
  I_c(p) &= D_c(p) + B_c(p) \label{eq:imagemodela}\\
    &= J_c(p) \cdot e^{-\beta_c^D \cdot d(p)} + B^{\infty}_c(p) \cdot (1-e^{-\beta^B_c \cdot d(p)}),  \label{eq:imagemodelb} 
\end{align}
\end{subequations}
where $c\in\{R,G,B\}$ is the color channel, $p$ is the pixel position, $I\in \mathbb{R}^{3\times H\times W}$ is the observed underwater image, $J\in \mathbb{R}^{3\times H\times W}$ is the restored clear image (i.e., scene radiance), and $B^{\infty}\in \mathbb{R}^{3\times H\times W}$ is the veiling light (i.e., global background light). $I_c(p)$ represents the pixel value of the $c$ channel at pixel position $p$ in image $I$. $\beta^D_c, \beta^B_c\in\mathbb{R}$ are the attenuation and scattering coefficients of the $c$ channel. $d\in \mathbb{R}^{1\times H\times W}$ is the per-pixel scene depth, and $d(p)$ represents the distance from the camera to the corresponding object point $p$ in the scene. As shown on the right side of Fig.~\ref{fig:physicalmodel}, red light is absorbed the most in water, resulting in underwater images typically having a bluish-green hue. In some previous literature~\cite{drews2013transmission,galdran2015automatic,fu2022unsupervised,cong2023pugan}, it was assumed that $\beta^D=\beta^B=\beta$. Therefore, the underwater physical imaging model can be rewritten as:
\begin{equation}
\label{eq:imagemodel1}
  I_c(p) = J_c(p) \cdot T_c(p) + B^{\infty}_c(p) \cdot (1-T_c(p)),   
\end{equation}
where $T_c(p)=e^{-\beta_c \cdot d(p)}$ and $T$ is called the transmission map. In this paper, we consider distinguishing between $\beta^D$ and $\beta^B$ to achieve more accurate modeling. Besides, the veiling light $B^{\infty}$ in Eq. (\ref{eq:imagemodelb}) is assumed to be spatially uniform, implying that it consists of a single and uniform color across the entire image. But this assumption is not reasonable in deep-sea scenarios with artificial light sources where the lighting is typically uneven. In this paper, we address this by assuming that the veiling light may vary at different pixel positions to adapt to scenes with uneven artificial lighting.


\subsection{Underwater Image Enhancement}
Underwater image enhancement methods can be categorized into three types: physical model-free, physical model-based, and deep learning-based methods. These methods are mainly applied in shallow-water environments and perform poorly in deep-sea environments. Therefore, some methods designed for deep-sea environments have been proposed recently.

\paragraph{Physical model-free methods}
Physical model-free methods directly adjust image pixel values based on various metrics~\cite{hitam2013mixture}. To improve color cast and contrast degradation in underwater images, some methods perform weighted summation of multiple enhanced versions~\cite{zhang2023underwater}. For example, Fusion~\cite{fusion} combines the results of white balance and contrast local adaptive histogram equalization. ACCC~\cite{accc} integrates the complementary advantages of local and global contrast-enhanced versions through multi-scale fusion~\cite{fusionv2}. MLLE~\cite{mlle} explores adaptive algorithms for color correction and contrast enhancement. Additionally, some methods based on Retinex theory decompose underwater images into reflection and illumination components and enhance them separately~\cite{fu2014retinex,zhang2017underwater}. However, these methods often suffer from over-enhancement and
color distortions in complex underwater environments.

\paragraph{Physical model-based methods}
Physical model-based methods estimate the parameters of the underwater physical imaging model in Eq. (\ref{eq:imagemodelb}) or (\ref{eq:imagemodel1}) to invert the degradation process. These parameters can be estimated based on various priors, such as the dark channel prior (DCP)~\cite{galdran2015automatic}, maximum intensity prior~\cite{carlevaris2010initial}, image blurriness~\cite{peng2015single,peng2017underwater}, and minimum information loss~\cite{li2016underwater}. UDCP~\cite{drews2013transmission,drews2016underwater} adapts DCP to underwater environments by using only the blue and green channels to estimate the transmission map. Sea-thru~\cite{akkaynak2019sea} proposes a physically accurate model and restores color based on RGBD images. Berman et al.~\cite{berman2020underwater} estimate the attenuation ratios of the blue-red and blue-green color channels by evaluating every possible water type. Zhou et al.~\cite{zhou2023underwater} estimate depth maps using a channel intensity prior and eliminate backscatter through adaptive dark pixels. However, these methods are sensitive to the assumptions made during parameter estimation, making them less robust in complex underwater environments.

\paragraph{Deep learning-based methods}
Deep learning-based methods have gained popularity recently due to their remarkable performance~\cite{rao2023deep}. Li et al.~\cite{uieb} introduce the UIEB dataset, which consists of many underwater images and corresponding reference images, for supervised training. This dataset has facilitated the development of various advanced UIE models~\cite{uiec,fivenet,cheng2024fdce}. Peng et al.~\cite{utrans} construct a lager dataset LSUI and design a Transformer-based UIE model. In addition, UGAN~\cite{ugan} and FUnIEGAN~\cite{funiegan} utilize adversarial training to generate clean images. Similar to our work, some methods integrate physical imaging models with neural networks. For instance, some researchers incorporate transmission maps and depth maps as attention or additional inputs to guide neural networks~\cite{skinner2019uwstereonet,ucolor}. PUGAN~\cite{cong2023pugan} estimates attenuation coefficients and depth maps to derive transmission maps, which then guide the decoding process of recovered images. To circumvent the reliance on paired training data, several self-supervised methods~\cite{kar2021zero,fu2022unsupervised,varghese2023self,yan2023hybrur} employ neural networks to estimate transmission maps, global background light, and even depth maps, scattering coefficients. These parameters are subsequently used to construct inversion models for image recovery. In comparison to these methods, our approach enables more accurate parameter estimation and better adaptation to various complex underwater environments.

\begin{figure*}[t]
\centering
\includegraphics[width=6in]{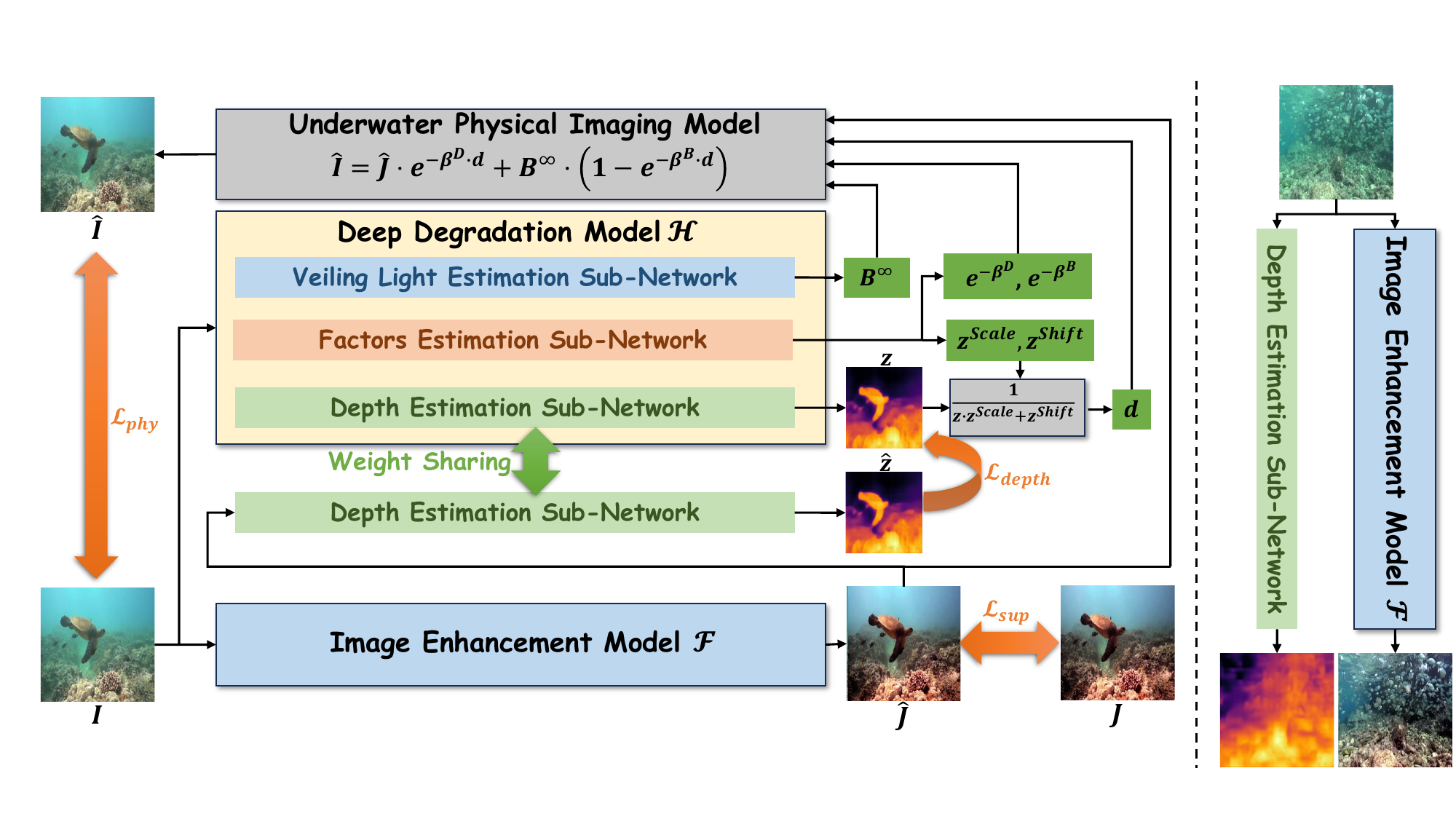}
\caption{Schematic diagram of our physical model-guided framework. It primarily consists of a deep degradation model (DDM) $\mathcal H$ and an underwater image enhancement (UIE) model $\mathcal{F}$. The UIE Model is used to obtain the enhanced image $\hat{J}$. The DDM is composed of three sub-networks that estimate various parameters of the imaging model. Using the outputs from $\mathcal H$ and $\mathcal F$, the re-degraded image $\hat{I}$ can be generated by the underwater physical imaging model. For simplicity, we omit the channel $c$ and pixel position $p$ in the equation.}
\vspace{-0.1cm}
\label{fig:framework}
\end{figure*}

\paragraph{Deep-sea image enhancement methods}
Underwater images in deep-sea environments often suffer from uneven lighting and low light conditions. L$^2$UWE~\cite{marques2020l2uwe} proposes two contrast-guided atmospheric
illumination models that enhance details and reduce dark regions. Cao et al.~\cite{cao2020nuicnet} simulate point light sources by adding light spots to the ground truth, creating a dataset that includes both synthetic and real data for training neural networks. Hou et al.~\cite{hou2023non} observe that the illumination channel of a uniform-light underwater image in the HSI color space contains few pixels close to zero, and present an effective illumination
channel sparsity prior (ICSP) based on this observation. IACC~\cite{zhou2024iacc} unifies the luminance features of underwater artificial and natural light and guides consistent enhancement across similar luminance regions. Other methods use CNNs to enhance non-uniform illumination images from the HSV~\cite{zhang2023framework} and LAB~\cite{li2023uialn} color spaces.

\subsection{Monocular Depth Estimation}

Recently, various monocular depth estimation (MDE) algorithms~\cite{ranftl2020towards,depthanything} have been proposed to estimate scene depth from a single image. These methods typically require training on datasets with detailed depth annotations~\cite{kitti}. However, collecting labeled datasets in underwater scenes is more challenging. Therefore, some researchers synthesize underwater-styled images by processing in-air images with depth data and physical imaging models to construct labeled datasets~\cite{uwcnn,hambarde2021uw}. Nonetheless, a domain gap exists between synthetic and real underwater images. To address this issue, Atlantis~\cite{atlantis} proposes a novel pipeline for generating photorealistic underwater images based on terrestrial depth data and diffusion models. Additionally, some works attempt to estimate depth in a self-supervised manner using GANs~\cite{gupta2019unsupervised} or the relationships between consecutive frames in underwater videos~\cite{wang2023underwater,varghese2023self}. Many physical model-based UIE algorithms indirectly obtain depth maps while estimating the transmission maps~\cite{drews2013transmission,galdran2015automatic,drews2016underwater,peng2017underwater,peng2018generalization}. However, due to unknown scattering and attenuation coefficients, the estimated depth maps are most likely wrong. A recent work, Depth Anything~\cite{depthanything}, demonstrates strong performance across various complex scenarios due to its pre-training on large-scale datasets. Although it is not specifically designed for underwater scenes, it serves as a good starting point for estimating depth in underwater environments.

\section{Methodology}
\subsection{Overview of Framework}

Our proposed framework is illustrated in Fig.~\ref{fig:framework}. Given an underwater image \(I\), the UIE model \(\mathcal{F}\) produces an enhanced image \(\hat{J}\). The supervised loss $\mathcal{L}_{sup}$ ensures that \(\hat{J}\) closely approximates the reference image \(J\). Additionally, \(I\) is input into a Deep Degradation Model (DDM) \(\mathcal{H}\), which comprises three sub-networks: veiling light estimation sub-network (VLEN), depth estimation network (DEN), and factors estimation sub-network (FEN). VLEN and DEN 
estimate the veiling light $B^{\infty}$ and relative inverse depth map $z$ of the underwater image \(I\), respectively. FEN estimates the attenuation coefficient, scattering coefficient, and the scale factor $z^{Sclae}$ and shift factor $z^{Shift}$ for the relative inverse depth. The two factors $z^{Sclae}$, $z^{Shift}$, and relative inverse depth map $z$ together derive the scene depth map \(d\). Finally, all variables are substituted into the underwater physical imaging model in Eq. (\ref{eq:imagemodelb}), resulting in a re-degraded image \(\hat{I}\). We enforce consistency between \(I\) and \(\hat{I}\) to impose a physical constraint loss $\mathcal{L}_{phy}$. It is noteworthy that for the parameters of the underwater physical imaging model output by DDM, such as scene depth $d$ and scattering coefficients $\beta^B$, we do not have their ground-truth values for supervised training. Therefore, we ingeniously utilize the physical constraint loss $\mathcal{L}_{phy}$, derived from the relationship between the raw and re-degraded images, to guide the DDM toward accurate estimations. In turn, these estimated parameters provide physical guidance for the training of the UIE model. Besides, considering that scene depth should be consistent before and after enhancement, we also utilize DEN to estimate the inverse depth map $\hat z$ of the enhanced image \(\hat{J}\) and introduce an additional depth consistency loss $\mathcal{L}_{depth}$. Ultimately, three losses jointly train the learnable components \(\mathcal{H}\) and \(\mathcal{F}\) within our framework. We will detail the structures of \(\mathcal{H}\) and \(\mathcal{F}\) in the following subsections.


\subsection{Deep Degradation Model}

\begin{figure}[t]
\centering
\includegraphics[width=3.49in]{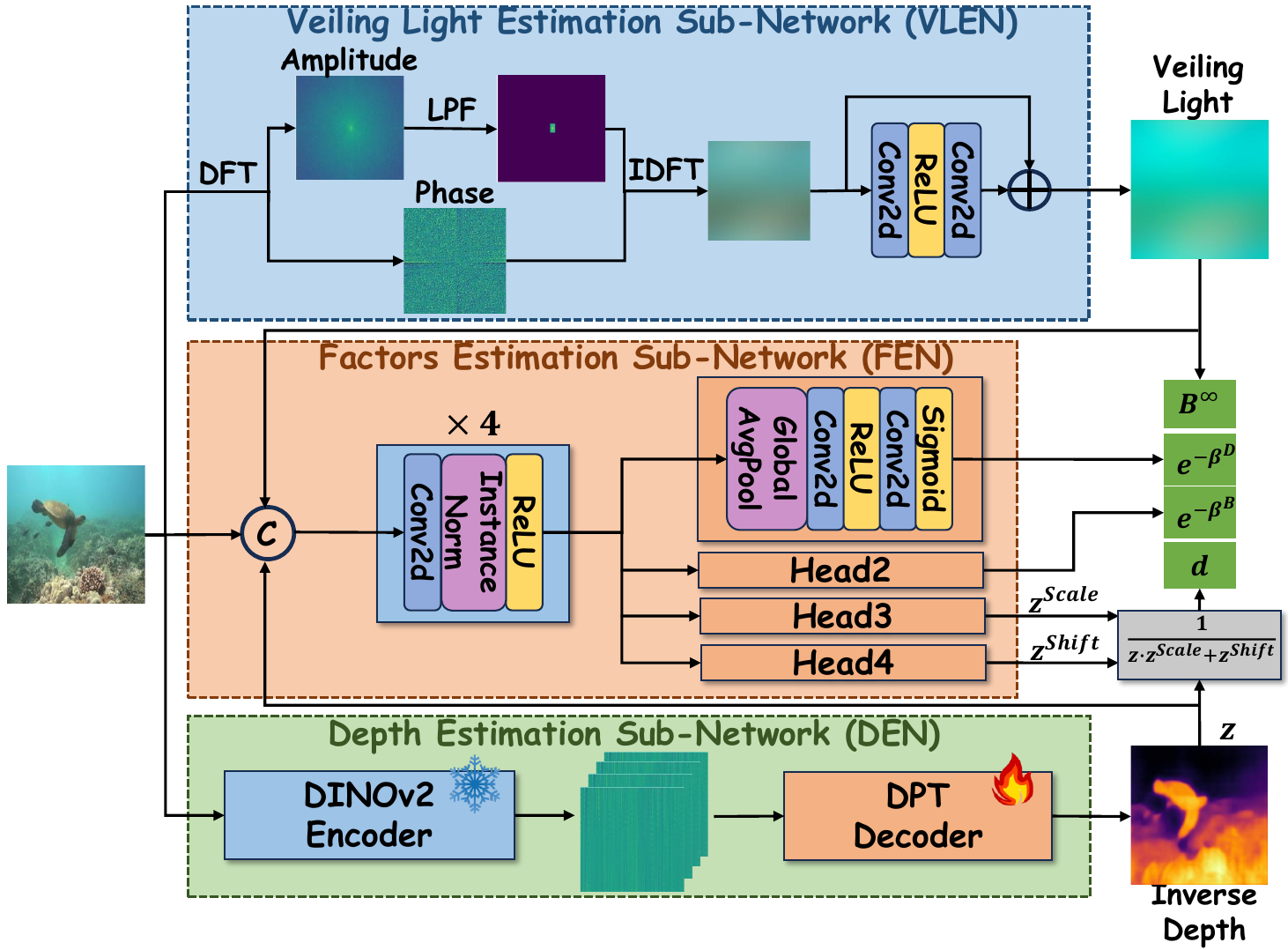}
\caption{Detailed structure of the three sub-networks in the deep degradation model, where $B^{\infty}$ represents the veiling light map, $e^{-\beta^D}$ and $e^{-\beta^B}$ denote the attenuation and scattering coefficients, $z$ indicates the disparity, while $z^{Scale}$ and $z^{Shift}$ are the scale and shift factors for depth transformation. The metric depth $d$ is calculated from $z$, $z^{Scale}$ and $z^{Shift}$.}
\label{fig:ddm}
\end{figure}

As shown in Fig.~\ref{fig:ddm}, the deep degradation model (DDM) $\mathcal{H}$ consists of three well-designed sub-networks that accurately estimate various parameters of the underwater physical imaging model in Eq. (\ref{eq:imagemodelb}). Below, we provide a detailed explanation of the design principles and structures.

\paragraph{Veiling Light Estimation Sub-Network} 

\begin{figure}[ht]
\centering
\includegraphics[width=2.7in]{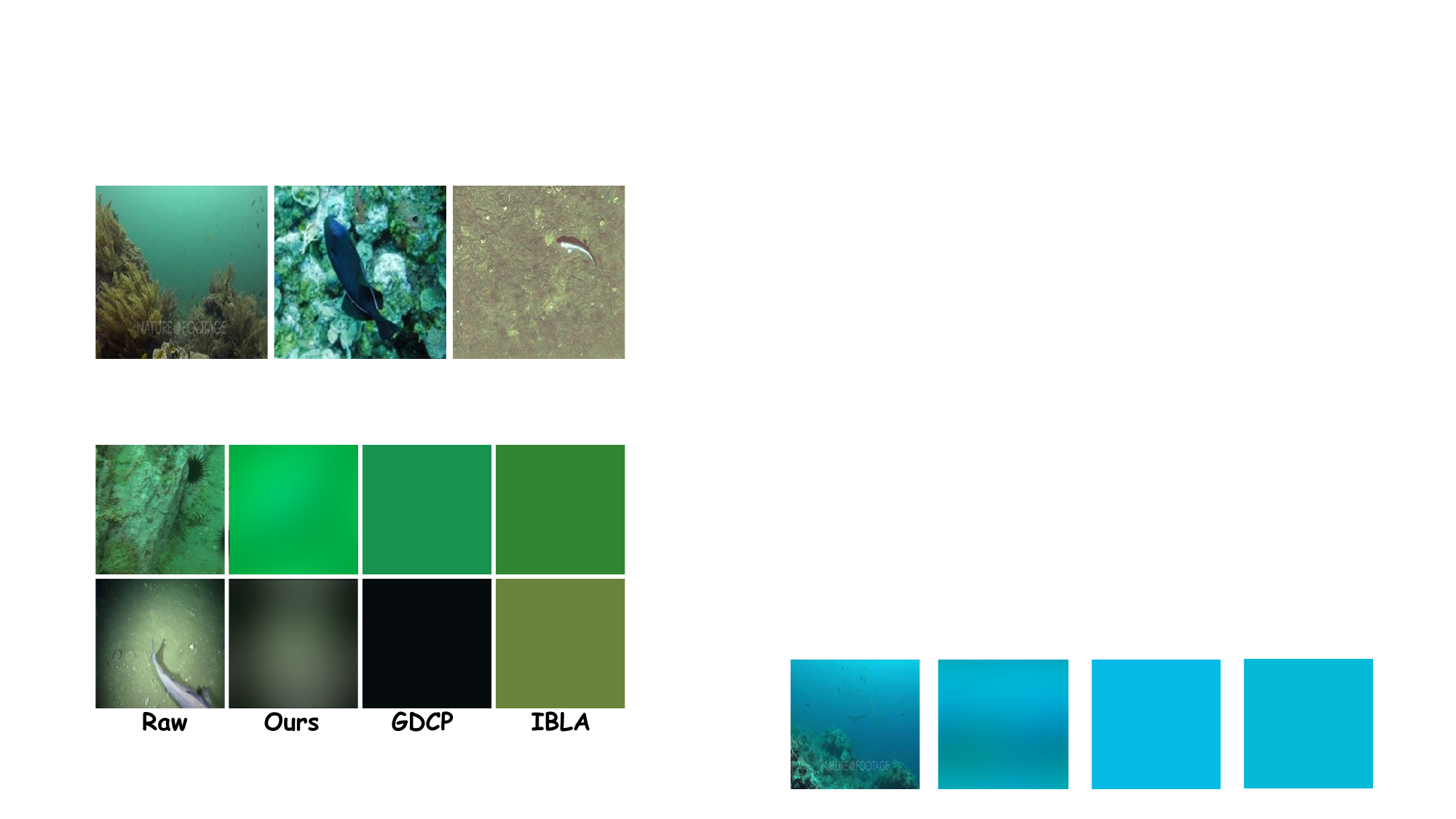}
\caption{Two underwater images under different lighting conditions and the veiling light maps estimated by three methods.}
\label{fig:differentlight}
\end{figure}

Most previous works assume that the veiling light $B^{\infty}$ can be represented by a uniform color image with identical pixel values across spatial locations. However, this assumption fails in deep-sea scenarios where the light primarily comes from artificial point light sources. As shown in the two examples on the far right of Fig.~\ref{fig:examples}, deep-sea images suffer from non-uniform illumination phenomena. Therefore, it is necessary to design a more reasonable veiling light estimation method to adapt to various lighting conditions from shallow to deep-sea environments. Fourmer~\cite{zhou2023fourmer} demonstrated that brightness, as a global feature, is primarily preserved in the center of the amplitude component of the image. Inspired by this idea, we first perform a Discrete Fourier Transform (DFT) on the underwater image to obtain the amplitude and phase components. Then we obtain the initial veiling light estimation by performing an Inverse DFT (IDFT) on the phase component and the amplitude component after applying a low-pass filter (LPF). To introduce flexibility and adjustability, the initial estimation is fed into convolutional modules with a residual connection to obtain the final veiling light $B^{\infty}$. Fig.~\ref{fig:differentlight} compares our method with IBLA~\cite{peng2017underwater} and GDCP~\cite{peng2018generalization}, both of which represent veiling light as uniform color images. For the first underwater image, the estimated veiling light maps of the three methods are similar. However, for the non-uniform illumination image in deep-sea scenarios, the other two methods fail while our method achieves relatively accurate veiling light estimation.

\begin{figure*}[t]
\centering
\includegraphics[width=6.1in]{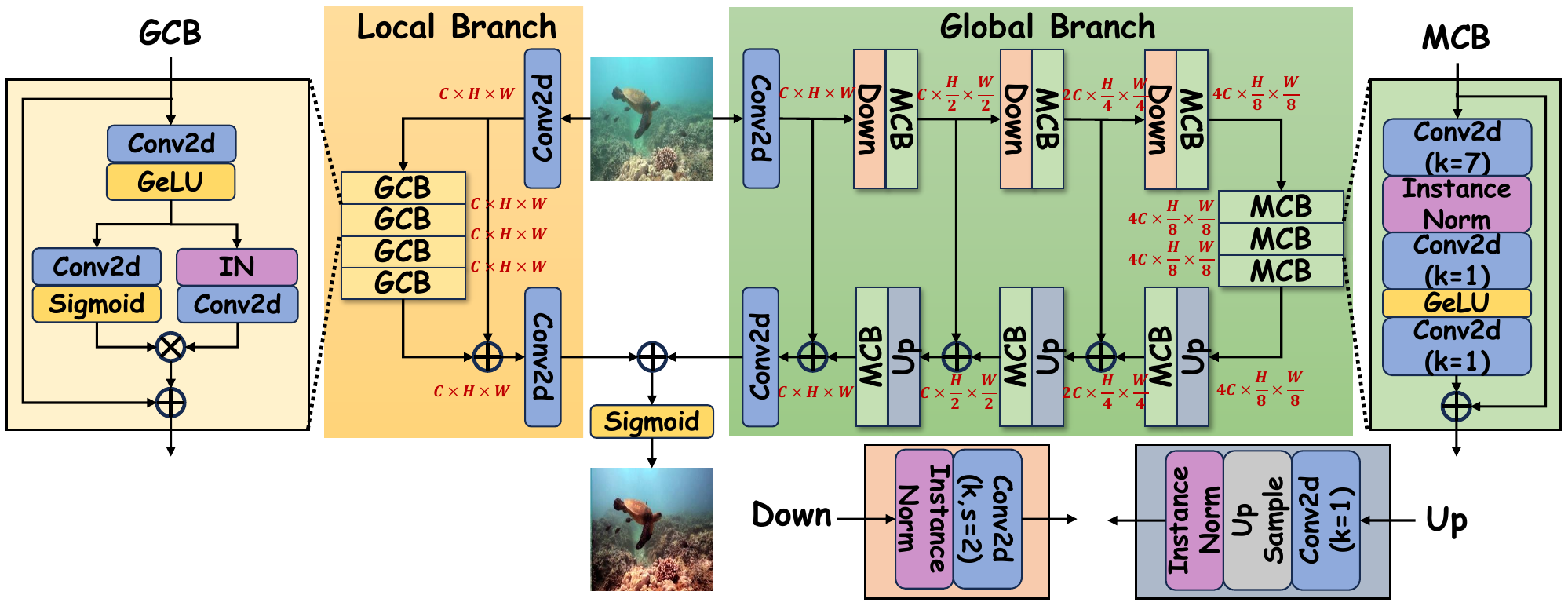}
\caption{The structural diagram of UIEConv. UIEConv includes the local branch on the left and the global branch on the right. We annotate the shapes of the intermediate features, where $C$, $H$, and $W$ represent the number of channels, height, and width. $k$ and $s$ represent the kernel size and stride in Conv2d.}
\label{fig:uieconv}
\end{figure*}

\begin{figure}[!htbp]
\centering
\includegraphics[width=2.7in]{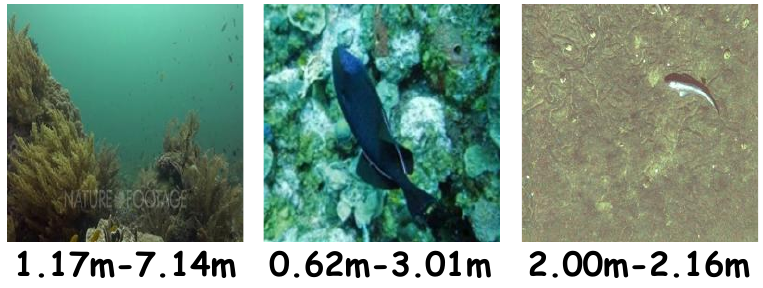}
\caption{Three underwater images with varying ranges of scene depth. The depth range below each image is derived from the output $z^{Scale}$ and $z^{Shift}$ of FEN. The first image features a large maximum depth in the water region; the second image has smaller maximum and minimum depths; in the last image, the minimum and maximum depths are nearly identical.}
\label{fig:differentdepth}
\end{figure}

\paragraph{Depth Estimation Sub-Network}

Scene depth is crucial for the physical imaging model, but accurately estimating it from a single underwater image is very challenging. A recent work, Depth Anything~\cite{depthanything}, demonstrates strong performance across various complex scenarios, including underwater environments, due to its pre-training on large-scale datasets. Depth Anything includes a DINOv2 encoder~\cite{dinov2} to extract image features and a dense prediction Transformer (DPT) decoder~\cite{dpt} to predict disparity, i.e., relative inverse depth. We aim to transfer the powerful zero-shot depth estimation capability of Depth Anything to our depth estimation sub-network (DEN). Specifically, we adopt the same encoder-decoder architecture and set the Dinov2 encoder to the small version of ViT~\cite{vit} for efficiency. We initialize DEN with the weights from Depth Anything. During training, we freeze the DINOv2 encoder weights and only fine-tune the DPT decoder. Ultimately, DEN can accurately estimate the normalized inverse depth map $z$.

\paragraph{Factors Estimation Sub-Network}

DEN estimates a relative inverse depth map, but what we need is the metric (absolute) depth map. To obtain the absolute depth $d$, previous methods used predefined maximum and minimum depth values, such as 0.1-6m, to scale the relative depth of all images to the same range. However, the depth ranges of different underwater images can vary significantly, as illustrated in Fig.~\ref{fig:differentdepth}. Therefore, we introduce a factors estimation sub-network (FEN) to estimate the scaling factor \(z^{Scale}\) and shift factor \(z^{Shift}\) for the relative inverse depth. The absolute depth \(d\) is then computed as \(d = \frac{1}{z\cdot z^{Scale} + z^{Shift}}\). Additionally, FEN is responsible for estimating the scattering and attenuation coefficients. Specifically, the underwater image $I$, the estimated veiling light map $B^{\infty}$, and the inverse depth map $z$ are concatenated along the channel dimension and then input into FEN, which consists of a backbone and four structurally identical yet independent heads. The backbone extracts general image degradation features, while the four independent heads specialize in mapping these general features to different parameters to be estimated, thereby achieving a decoupled estimation. Specifically, the backbone includes four simple convolutional layers, each comprising convolution, instance normalization, and a ReLU activation function. The four independent heads are designed to estimate \(z^{Scale}\), \(z^{Shift}\), \(e^{-\beta^D}\), and \(e^{-\beta^B}\), respectively. Given that the attenuation coefficient $\beta^D$ and scattering coefficient $\beta^B$ are positive values, a Sigmoid activation function is applied to produce values between 0 and 1 as the terms \(e^{-\beta^D}\) and \(e^{-\beta^B}\). To more reasonably scale the relative depth, we add 0.1 to the Sigmoid output to generate values greater than 0.1 as \(z^{Shift}\) and multiply the Sigmoid output by 2 to generate values between 0 and 2 as \(z^{Scale}\). The approach allows the maximum and minimum absolute depths to vary significantly within a range of 0-10m.

\subsection{UIEConv Model}

Although the enhancement model $\mathcal{F}$ in our framework can be any advanced UIE model, such as UShape~\cite{utrans} or UIEC$^2$Net~\cite{uiec}, we have designed a simple yet more effective fully convolutional model called UIEConv. As shown in Fig.~\ref{fig:uieconv}, UIEConv includes two branches: a global and a local branch.

\paragraph{Global Branch}
The global branch follows the common U-Netencoder-decoder architecture. In the encoding part, downsampling reduces the image resolution, while in the decoding part, upsampling gradually restores the image to its original resolution. We use convolutions with a stride and kernel size of 2 for downsampling and bilinear interpolation for upsampling. The core module of the global branch is the Modern Convolutional Block (MCB) derived from the advanced vision convolutional backbone ConvNeXt~\cite{convnext}, which employs large kernel convolutions (kernel size of 7) and a transformer-like~\cite{vit} design. We replace Layer Normalization~\cite{layernorm} with Instance Normalization~\cite{instancenorm}, which has been proven to be more suitable for image restoration tasks~\cite{du2024end}. The combination of the U-Net-like architecture and large kernel convolutions results in a very large receptive field, allowing the capture of global features in the image without the need for complex attention mechanisms.

\paragraph{Local Branch}
During the downsampling process in the global branch, the image resolution is reduced, leading to a loss of details. To focus on local areas and supplement detailed information, the local branch stacks four Gated Convolutional Blocks (GCB), maintaining resolution throughout the forward process. GCB primarily employs convolutions with a kernel size of 3 and GeLU nonlinear activation functions to extract local features. Considering that underwater images may have varying degrees of degradation across different spatial locations and channels, we also introduce gating mechanism~\cite{zhou2024iacc} in GCB. Specifically, we use convolutions and sigmoid activation functions to generate dynamic weights, which adaptively control the retention of features in different spatial locations and channels. The local branch, with its smaller receptive field, focuses on local features and image details.

The outputs of the global and local branches are added and passed through a Sigmoid activation function to obtain the final enhanced image $\hat J$.

\subsection{Loss Function}
The total loss function comprises three components: supervised loss $\mathcal{L}_{sup}$, physical constraint loss $\mathcal{L}_{phy}$, and depth consistency loss $\mathcal{L}_{depth}$. 

\paragraph{Supervised Loss}
The supervised loss calculates the L1 distance between the enhanced image $\hat J$ and the reference image $J$:
\begin{equation}
\label{eq:loss1}
  \mathcal{L}_{sup} = \frac{1}{HW}\sum_{i}^{H}\sum_{j}^{W}|J_{ij}-\hat J_{ij}|,
\end{equation}
where $H$ represents the height and $W$ represents the width of the image, and $ij$ represents the pixel position.

\paragraph{Physical Constraint Loss}
The physical constraint loss computes the L1 distance between the degraded image $\hat I$, which is obtained by applying the underwater physical imaging model to the enhanced image $\hat J$, and the original underwater image $I$:
\begin{equation}
\label{eq:loss2}
  \mathcal{L}_{phy} = \frac{1}{HW}\sum_{i}^{H}\sum_{j}^{W}|I_{ij}-\hat I_{ij}|.
\end{equation}
The physical constraint loss is a primary optimization objective for UIE models in some self-supervised UIE methods~\cite{fu2022unsupervised,varghese2023self}, while it serves as an additional regularization term to assist the training of the UIE model $\mathcal F$ in our framework.

\begin{table*}[htbp]
\begin{center}
  \caption{Quantitative comparison of various UIE methods across four datasets primarily consisting of shallow-water images. The best and second-best results are highlighted in red and blue, respectively.}
  \label{tab:sota}%
    \begin{tabular}{c||c|c||c|c||c|c||c|c||c|c|c}
    \hline
     & \multicolumn{2}{c||}{LSUI}  & \multicolumn{2}{c||}{UIEB}     & \multicolumn{2}{c||}{C60}  & \multicolumn{2}{c||}{U45} & {\multirow{2}{*}{FLOPs$\downarrow$}} & {\multirow{2}{*}{\#Param.$\downarrow$}}& {\multirow{2}{*}{Time$\downarrow$}}\\
    \cline{1-9}
     Method & PSNR$\uparrow$  & SSIM$\uparrow$  & PSNR$\uparrow$  & SSIM$\uparrow$  & UIQM$\uparrow$  & UCIQE$\uparrow$ & UIQM$\uparrow$  & UCIQE$\uparrow$ & & &\\
    \hline
    \hline
    GDCP~\cite{peng2018generalization} & 13.4928 & 0.6852 & 14.3821 & 0.7360 & 2.1487 & 0.5873 & 2.2481 & 0.5937 & -& -&0.076s\\
    Fusion~\cite{fusion} & 18.6835 & 0.7923 & 23.0684 & 0.9194 & 2.6077 & \textcolor{red}{0.6123} & 2.9691 & \textcolor{red}{0.6393} & -& -& 0.056s\\
    IBLA~\cite{peng2017underwater} & 17.2079 & 0.7516 & 18.6182 & 0.7715 & 1.9790 & 0.5874 & 2.3578 & 0.5836 & -& -&2.934s\\
    UGAN~\cite{ugan} & 21.1094 & 0.8103 & 19.5773 & 0.8280 & \textcolor{blue}{2.8587} & 0.5485 & 3.1022 & 0.5670 & 18.15G & 54.40M&0.006s \\
    FUnIEGAN~\cite{funiegan} & 21.7107 & 0.7967 & 19.7440 & 0.8381 & \textcolor{red}{2.8725} & 0.5457 & 2.9190 & 0.5594 & 10.24G & 7.02M&0.002s\\
    Ucolor~\cite{ucolor} & 21.2919 & 0.8324 & 23.2240 & 0.9039 & 2.6610 & 0.5520 & \textcolor{blue}{3.2023} & 0.5796 & 443.85G & 157.42M&0.832s\\
    PUGAN~\cite{cong2023pugan} & 20.5350 & 0.8075 & 22.6060 & 0.8774 & 2.8460 & \textcolor{blue}{0.5998} & 3.1774 & \textcolor{blue}{0.6124} & 72.05G & 95.66M&0.021s\\
    UIEC$^2$Net~\cite{uiec} & 25.0757 & 0.8708 & 23.3631 & 0.9008 & 2.8481 & 0.5810 & \textcolor{red}{3.2157} & 0.5820 & 26.06G & 0.53M &0.041s\\
    Ushape~\cite{utrans} & 25.7630 & 0.8296 & 20.3947 & 0.7763 & 2.4828 & 0.5475 & 2.8956 & 0.5725 & 2.98G & 22.82M &0.046s\\
    FDCE-Net~\cite{cheng2024fdce} & 25.2461 & 0.8442 & 21.3483 & 0.8467 & 2.5812 & 0.5530 & 3.0121 & 0.5885 & 12.78G & 5.53M &0.025s\\
    UIR-PolyKernel~\cite{guo2025underwater} & 26.4282 & 0.8893 & 22.1323 & 0.8931 & 2.4031 & 0.5638 & 2.9458 & 0.5787 & 13.67G & 1.84M &0.029s\\
    \hline
    UIEConv & \textcolor{blue}{28.9155} & \textcolor{blue}{0.9187} & \textcolor{blue}{24.1197} & \textcolor{blue}{0.9283} & 2.3540 & 0.5709 & 3.0953 & 0.5907 & 121.53G & 3.31M & 0.020s\\
    Ours & \textcolor{red}{29.9253} & \textcolor{red}{0.9248} & \textcolor{red}{24.9395} & \textcolor{red}{0.9429} & 2.5486 & 0.5767 & 3.1542 & 0.5966 & 146.91G & 29.55M & 0.049s\\
    \hline
    \end{tabular}%
\end{center}
\end{table*}%

\begin{figure*}[!htbp]
\centering
\includegraphics[width=6.4in]{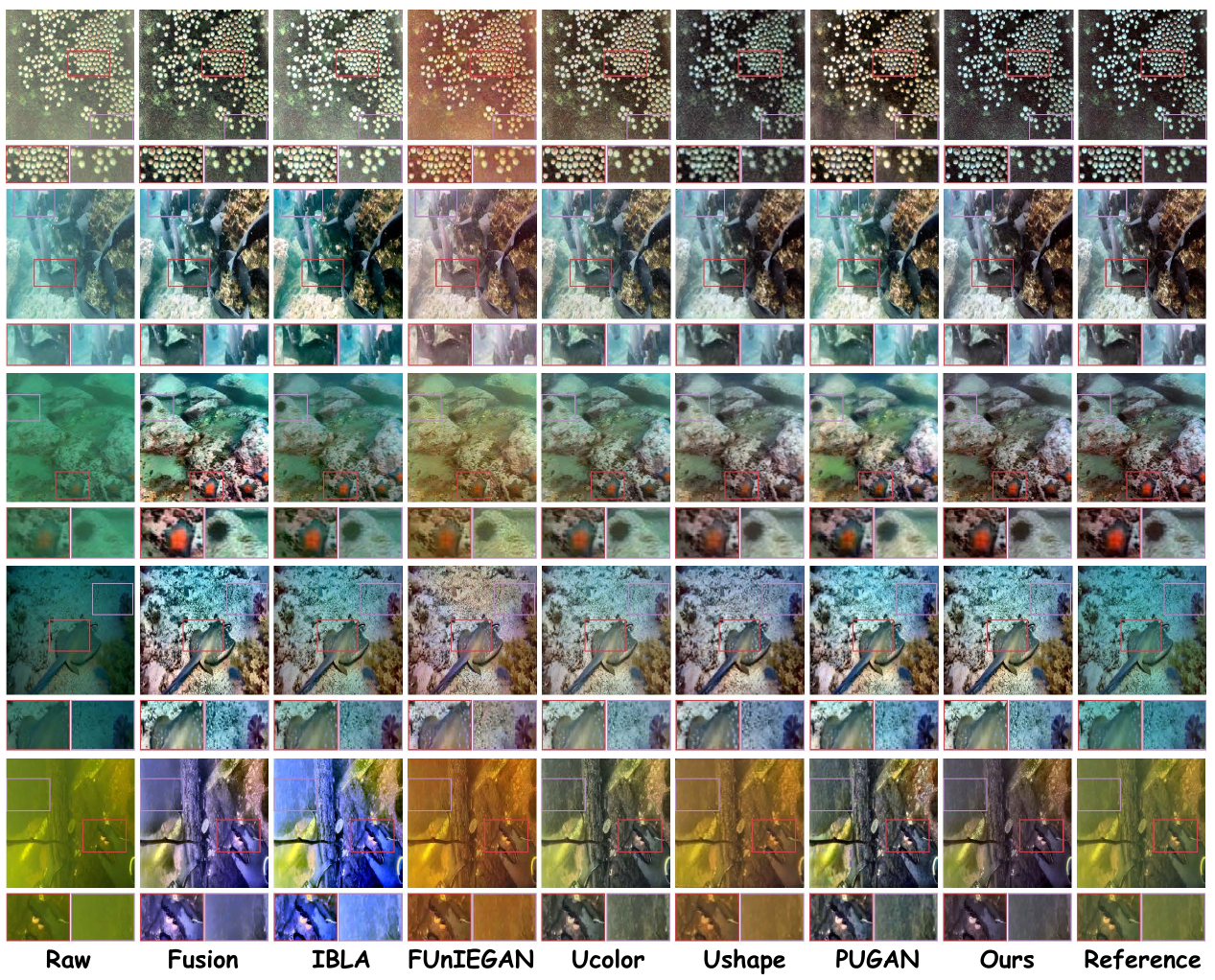}
\caption{Enhancement results of five images from the LSUI and UIEB dataset. The first three images come from LSUI, while the last two come from UIEB.}
\vspace{-0.1cm}
\label{fig:comparelusit90}
\end{figure*}

\paragraph{Depth Consistency Loss}
Enhancing an image should not alter the scene depth. A robust depth estimation sub-network should consistently output the same depth map regardless of image quality changes. Therefore, we compute the scale- and shift-invariant loss~\cite{ranftl2020towards} between the inverse depth map $z$  of the underwater image and the inverse depth map $\hat z$ of the enhanced image as the depth consistency loss:
\begin{equation}
\label{eq:loss3}
  \mathcal{L}_{depth} = \frac{1}{HW}\sum_{i}^{H}\sum_{j}^{W}\rho(z_{ij},\hat z_{ij}),
\end{equation}
where $\rho$ is the affine-invariant mean absolute error loss defined in~\cite{depthanything}.

The overall loss function is a weighted sum of the above three losses:
\begin{equation}
\label{eq:loss}
  \mathcal{L} = \mathcal{L}_{sup} + \lambda_1\mathcal{L}_{phy} + \lambda_2\mathcal{L}_{depth},
\end{equation}
where $\lambda_1$ and $\lambda_2$ are empirically set to 0.2 and 1, respectively.

\section{Experiments}
\subsection{Datasets and Evaluation Metrics}
We conduct experiments on three shallow-water datasets (where the light source for most images comes from sunlight) and two deep-sea datasets (where almost all images are illuminated by artificial lights). Additionally, we perform quantitative evaluations of depth estimation on an underwater dataset with depth annotations.

The shallow water datasets include LSUI~\cite{utrans}, UIEB~\cite{uieb} and U45~\cite{u45}. \textbf{LSUI} contains 4279 pairs of real-world underwater images and clear reference images. Following~\cite{utrans}, we use 3879 pairs of these images as the training set, and the remaining 400 pairs of images as the test set. To evaluate generalization, we use 890 images with corresponding high-quality reference images in \textbf{UIEB} as the second test set. In addition, UIEB also includes a set of 60 challenge images (\textbf{C60}) that do not have corresponding reference images. And \textbf{U45} contains 45 carefully selected underwater images, serving as an important benchmark for UIE. For all the supervised methods, we train the models on the training set of LSUI and test them on the LUSI test set, UIEB, C60, and U45. The deep-sea datasets are UIID~\cite{cao2020nuicnet} and OceanDark~\cite{oceandark}. \textbf{UIID} contains 3486 pairs of non-uniform illumination images with reference images. This dataset consists primarily of synthetic images, with a small portion of real images. We randomly selecte 3136 pairs for the training set and reserve the remaining 350 pairs for the test set. \textbf{OceanDark} comprises 183 underwater images captured by video cameras located in profound depths using artificial lighting. For all supervised methods, we train the models on the UIID training set and evaluate them on both the UIID test set and OceanDark. To quantitatively evaluate depth estimation, we conduct evaluations using the Sea-thru’s \textbf{D3} and \textbf{D5} subsets~\cite{akkaynak2019sea}, which includes underwater images and corresponding depth maps obtained via the Structure-from-Motion algorithm.

We use the commonly employed PSNR and SSIM as full-reference image quality evaluation metrics. For datasets without reference images, we use UIQM~\cite{UIQM} and UCIQE~\cite{UCIQE}, which are designed for underwater image quality assessment, as no-reference metrics. In deep-sea scenarios, we also use the PIQE~\cite{PIQE} metric. To compare the efficiency of different methods, we report the runtime for all methods and the parameter count and FLOPs only for the deep learning-based models. For depth estimation, we use six metrics in~\cite{depthanything}: root mean square
error (RMSE), absolute mean relative error (Abs.Rel), absolute error in log-scale (log$_{10}$), and the percentage of inlier pixels ($\delta_i$) with threshold $1.25^i$.

\subsection{Implementation Details}

We replicated all the compared methods based on their official codes and hyper-parameters. All models were trained for 160 epochs with a batch size of 8, using the ADAM optimizer. The initial learning rate was set to $5 \times 10^{-5}$ and reduced by half after 128 epochs. In our proposed training framework, the learning rate for the depth estimation sub-network was set to 0.3 times that of the other modules. To ensure a fair comparison, all images were resized to 256$\times$256 during both training and testing. For runtime evaluation, we ran all deep learning-based models on an NVIDIA TITAN V GPU and traditional methods on an Intel(R) Core(TM) i7-13700K CPU. We inferenced on 1000 images to report the average runtime per image.

\subsection{Performance Evaluation}
We comprehensively compare our method with other UIE methods in various scenarios. Besides, we qualitatively and quantitatively evaluate the performance of depth estimation.

\begin{figure*}[!htbp]
\centering
\includegraphics[width=6.4in]{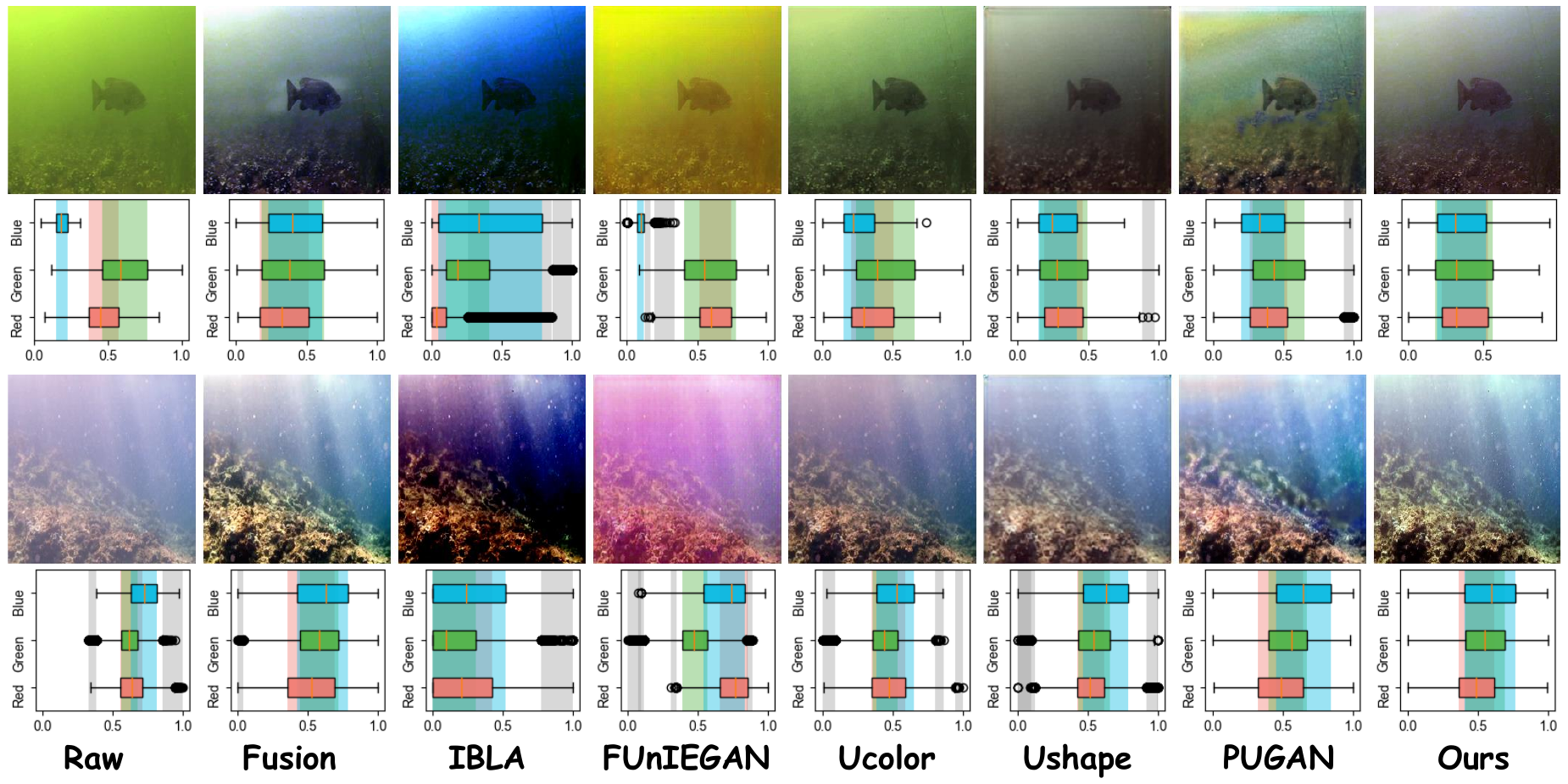}
\caption{Enhancement results of two images from the C60 dataset. The intensity distribution of the RGB channels is presented in horizontally placed box plots which intuitively display the upper edge, upper quartile, median, lower quartile, and lower edge of each channel. The black points represent outliers in each channel, and the gray rectangles indicate the areas where outliers are clustered.}
\label{fig:comparec60}
\end{figure*}

\begin{figure*}[!htbp]
\centering
\includegraphics[width=6.4in]{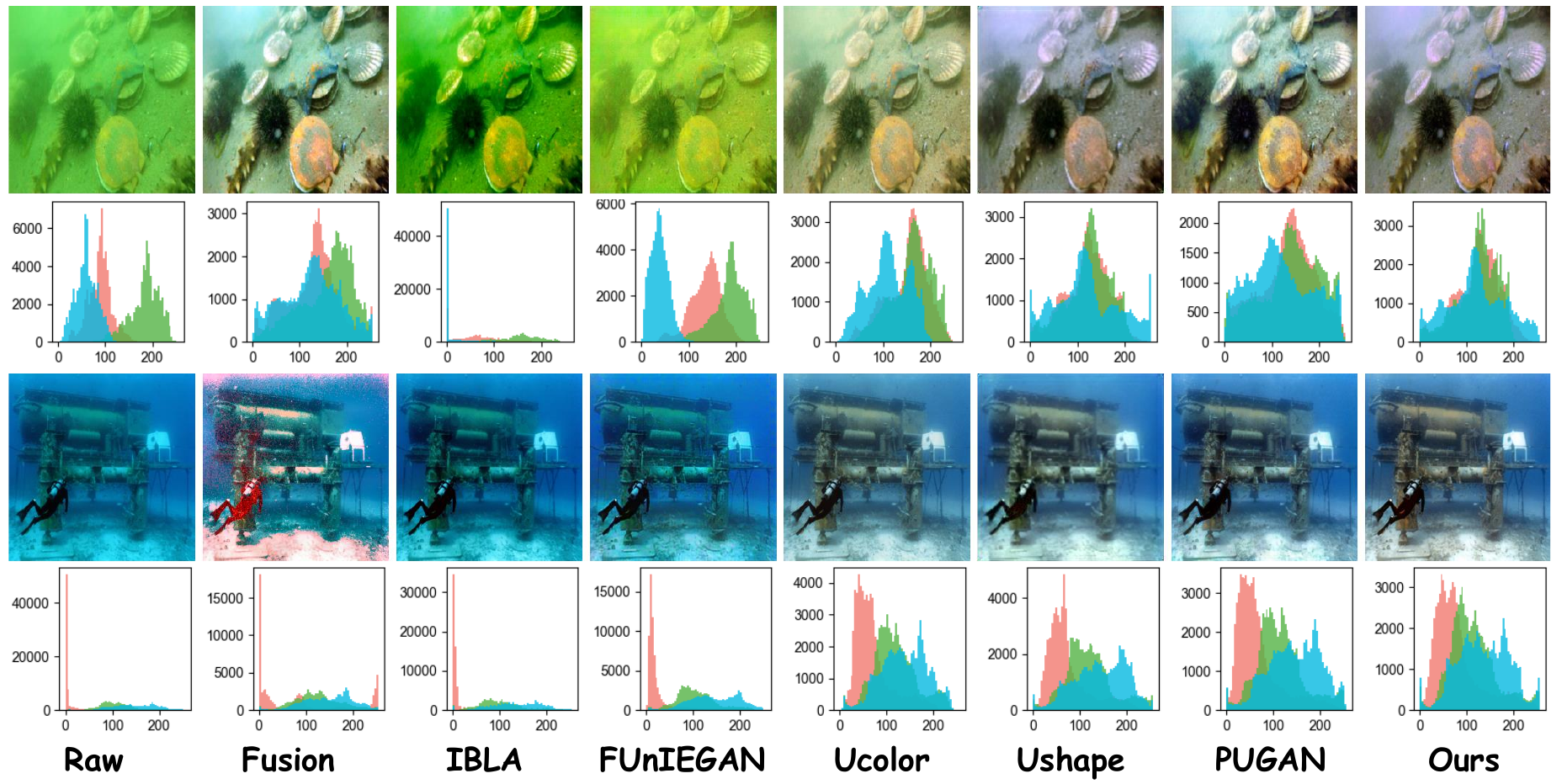}
\caption{Enhancement results of two images from the U45 dataset. We display the histogram distribution of the RGB channels.}
\label{fig:compareu45}
\end{figure*}

\paragraph{Shallow-water Scene}

We use the training set of LSUI for training and the LSUI test set, UIEB, C60, and U45 for testing. Almost all the images in these datasets come from shallow-sea scenes, where the light mainly come from the sun. In Table~\ref{tab:sota}, we report not only the results of training UIEConv integrated into our framework (last row) but also the results of training UIEConv alone (second to last row). For datasets with reference images like LSUI and UIEB, our method achieves the best PSNR and SSIM metrics. For images without reference images in C60 and U45, Fusion achieves the best UCIQE, while FUnIEGAN and UIEC2Net perform best on UIQM. However, as shown in Figs.~\ref{fig:comparelusit90},~\ref{fig:comparec60},~\ref{fig:compareu45}, their enhancement effects are not satisfactory from a human visual perspective. This observation suggests that these two non-reference quality assessment metrics sometimes do not align with human visual perception, as noted in many previous works~\cite{uieb,cong2023pugan,du2024end}. In contrast, our model consistently produces enhanced images with no color cast and clear details across various underwater conditions and degradation effects. In the last two examples, our enhanced images even appear more satisfactory than the reference images, removing excessive blue and yellow tones. Additionally, the UIEConv model has a small number of parameters and runs quickly. While integrating UIEConv into our framework reduces efficiency, it significantly improves performance. Future work will focus on designing more efficient sub-network structures.

\begin{table*}[htbp]
\begin{center}
  \caption{Quantitative comparison of various UIE methods across two datasets featuring deep-sea images with uneven lighting. The best and second-best results are highlighted in red and blue, respectively.}
  \label{tab:sota_deepsea}%
    \begin{tabular}{c||c|c|c|c|c||c|c|c||c|c|c}
    \hline
     & \multicolumn{5}{c||}{UIID}             & \multicolumn{3}{c||}{OceanDark} & {\multirow{2}{*}{FLOPs$\downarrow$}} & {\multirow{2}{*}{\#Param.$\downarrow$}}& {\multirow{2}{*}{Time$\downarrow$}}\\
    \cline{1-9}
     Method & PSNR$\uparrow$  & SSIM$\uparrow$  & UIQM$\uparrow$  & UCIQE$\uparrow$ & PIQE$\downarrow$  & UIQM$\uparrow$  & UCIQE$\uparrow$ & PIQE$\downarrow$ & & &\\
    \hline
    \hline
    ICSP~\cite{hou2023non} & 8.6463 & 0.3544 & 1.1620 & \textcolor{red}{0.6251} & 20.4159 & 2.3674 & \textcolor{red}{0.5761} & 8.9793 & -& -&0.034s\\
    L$^{2}$UWE~\cite{marques2020l2uwe} & 11.1388 & 0.4964 & 1.9282 & \textcolor{blue}{0.6121} & 11.3622 & \textcolor{red}{3.2258} & \textcolor{blue}{0.5514} & 6.1632 & -& -&2.302s\\
    NUICNet~\cite{cao2020nuicnet} & 23.5712 & 0.8159 & 2.9312 & 0.5086 & \textcolor{red}{7.4141} & 2.6533 & 0.5081 & 5.6451 & 49.95G & 15.70M&0.014s\\
    IACC~\cite{zhou2024iacc}  & 25.0095 & 0.8895 & 3.0979 & 0.5137 & \textcolor{blue}{8.3629} & 2.8983 & 0.5148 & \textcolor{red}{5.1617} & 132.44G & 2.10M&0.054s\\
    \hline
    UIEConv & \textcolor{blue}{27.4934} & \textcolor{blue}{0.9129} & \textcolor{red}{3.1545} & 0.5231 & 8.5412 & \textcolor{blue}{2.9005} & 0.5094 & 5.5921 & 121.53G & 3.31M&0.020s\\
    Ours  & \textcolor{red}{28.6446} & \textcolor{red}{0.9203} & \textcolor{blue}{3.1226} & 0.5240 & 8.6548 & 2.8771 & 0.5059 & \textcolor{blue}{5.2309} & 146.91G & 29.55M&0.049s\\
    \hline
    \end{tabular}%
\end{center}
\end{table*}%

\begin{figure}[t]
\centering
\includegraphics[width=3.45in]{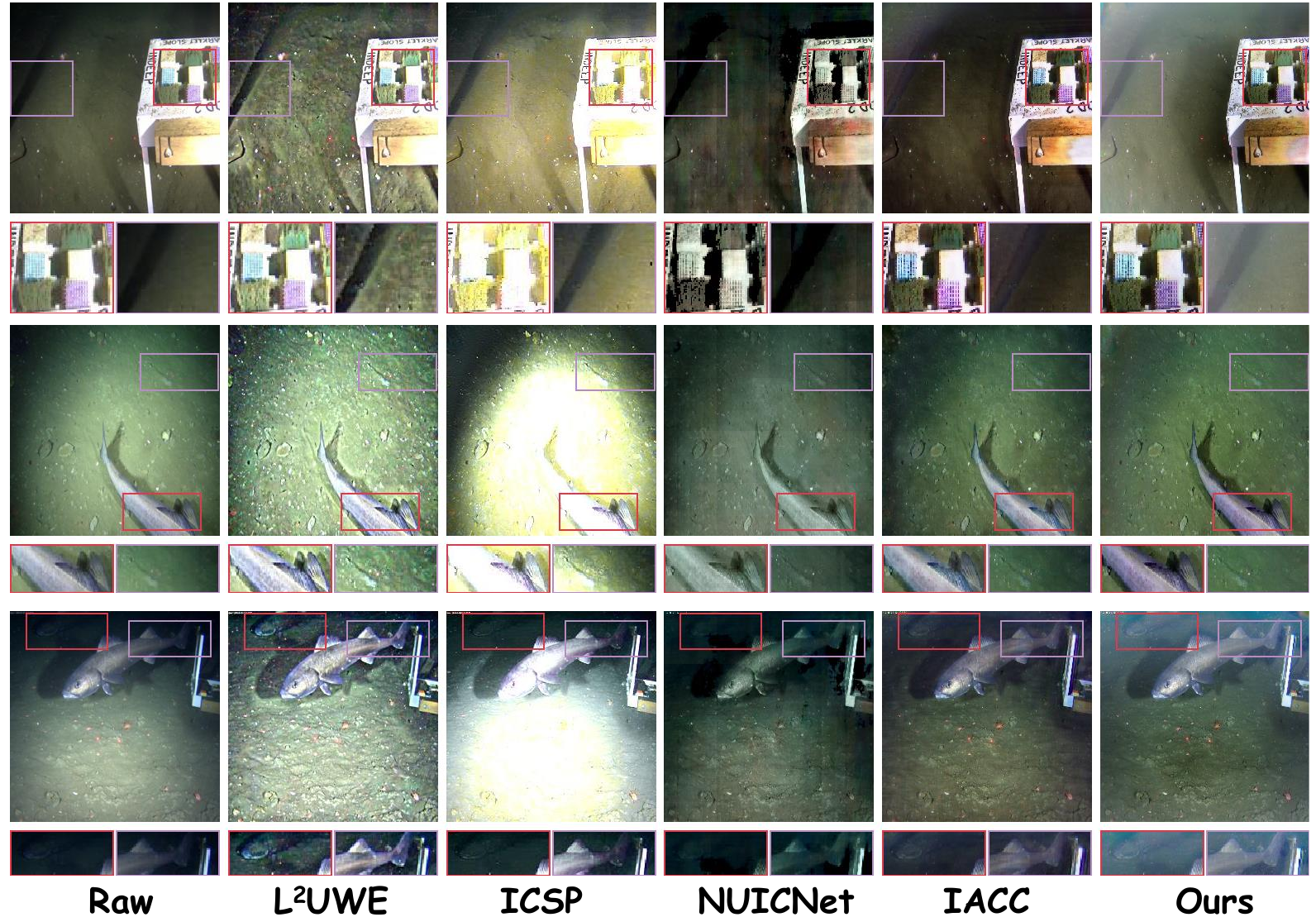}
\caption{Enhancement results of three deep-sea images with uneven lighting from the OceanDark dataset. We enlarge the local areas of the enhanced images to compare the details.}
\label{fig:compareoceandark}
\end{figure}

\paragraph{Challenging Scene} 
The C60 dataset consists of 60 challenging underwater images of very poor quality. As shown in Fig.~\ref{fig:comparec60}, two underwater images suffer from severe blurriness and abnormal colors, and other UIE methods perform poorly on these two images. For instance, the enhanced visual effect of IBLA is even worse than the original underwater image, and PUGAN's first enhanced image exhibits significant distortion. In contrast, our method achieves the best visual results. Moreover, the intensity distribution of the RGB channels in our enhanced images is more even and reasonable, with almost no outlier pixels.

\paragraph{Severe Color Distortion Scene} There are 45 images suffering from severe color casts in the U45 dataset. As shown in Fig.~\ref{fig:compareu45}, we present two typical examples: one image exhibits a pronounced green hue, while the other has a strong blue tint. Analyzing the RGB channel histograms reveals that the green channel histogram of the first image is skewed to the right, while the red channel in the second image is notably attenuated. Fusion overcompensates for the red channel, and IBLA fails to correct the color in both images. Our model demonstrates superior capability in suppressing dominant channels and compensating for weaker channels, thereby effectively correcting color casts. The color distribution of our enhanced images closely approximates that of in-air images, aligning well with the gray world assumption.

\paragraph{Deep-sea Scene} 
For deep-sea scenarios, we train on the UIID training set and test on the UIID test set and OceanDark dataset. Table~\ref{tab:sota_deepsea} compares our method with several UIE methods tailored for deep-sea environments with uneven lighting conditions. Our method achieves superior performance on two full-reference metrics, PSNR and SSIM. While some methods outperform ours on non-reference metrics, a visual comparison in Fig.~\ref{fig:compareoceandark} reveals the opposite. For instance, ICSP excels in the UCIQE metric but produces overexposed images. In contrast, our method enhances brightness in dark regions, mitigating uneven lighting without introducing distortion. This highlights the challenge of accurately evaluating deep-sea image quality using existing non-reference metrics. Developing appropriate non-reference metrics for deep-sea images remains an important research direction.

\begin{figure}[t]
\centering
\includegraphics[width=3.45in]{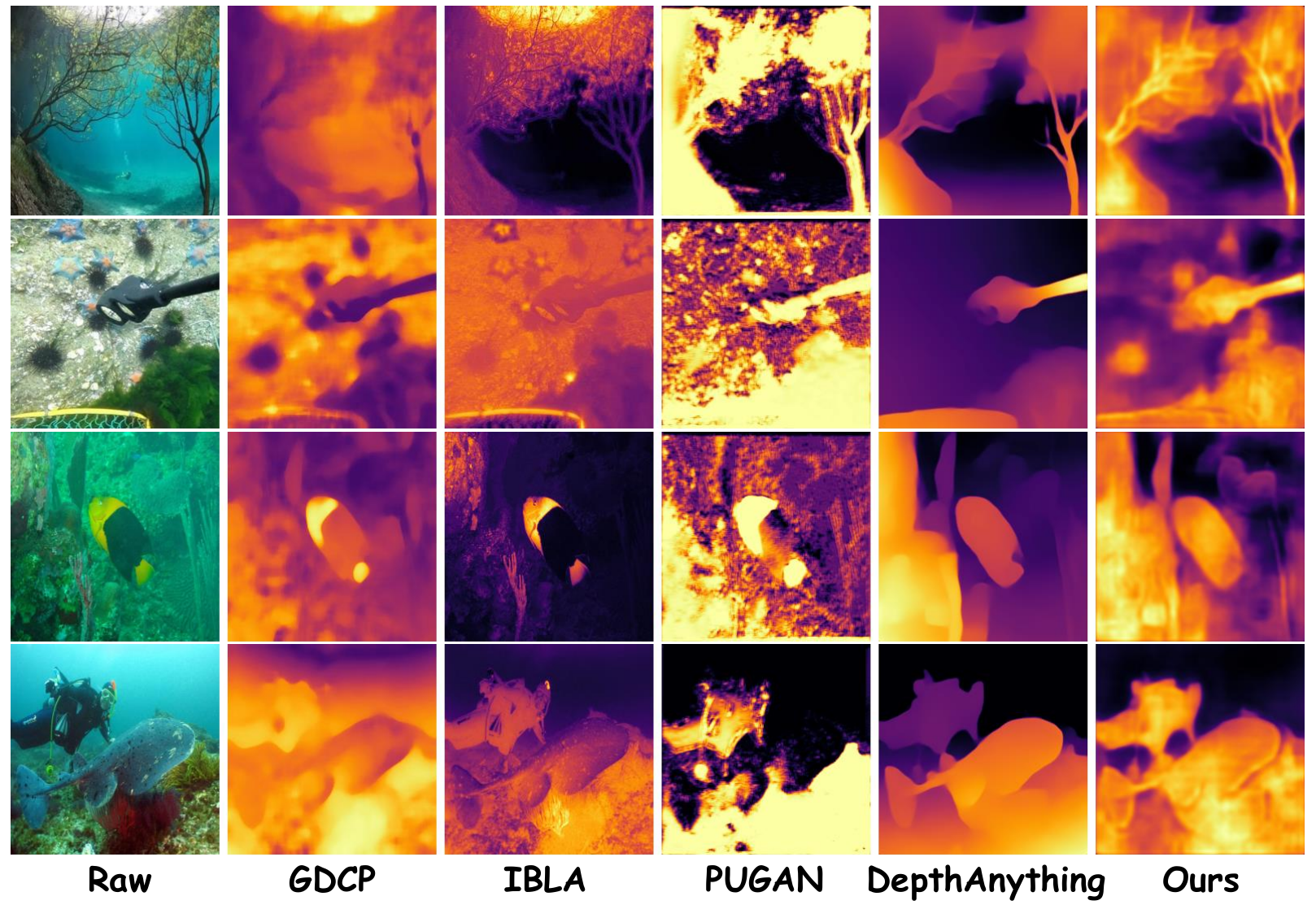}
\caption{Visual comparison of depth maps estimated by five different methods.}
\label{fig:comparedepth}
\end{figure}

\paragraph{Depth Estimation}
We visualize inverse depth maps estimated by several methods in Fig.~\ref{fig:comparedepth}. Methods like GDCP and IBLA, which use visual priors to estimate transmission and depth maps, perform the worst. PUGAN struggles in texture-rich scenes. Depth Anything exhibits strong generalization ability and adapts well to underwater scenes, but its depth maps are overly smooth. In comparison, our depth estimation sub-network produces more detailed depth maps, such as the branches in the first example and the sea urchins in the second example. Additionally, we quantitatively compare the performance of Depth Anything and our method on a real underwater image dataset with depth annotations in Table~\ref{tab:abla_depth}. Our method outperforms Depth Anything. Although our depth estimation sub-network is initialized from Depth Anything, through joint training with underwater image enhancement and physical imaging model parameter estimation, our method is better suited for underwater images.


\begin{table}[t]
\begin{center}
  \caption{Quantitative comparisons of depth estimation on real underwater images from two subsets of Sea-thru dataset. DA is an abbreviation for Depth Anything.}
  \label{tab:abla_depth}%
  \setlength{\tabcolsep}{5pt}
    \begin{tabular}{c||cccccc}
    \hline
     Method & RMSE$\downarrow$ & Abs.Rel$\downarrow$ & log$_{10}$$\downarrow$ & $\delta_1$$\uparrow$ & $\delta_2$$\uparrow$ & $\delta_3$$\uparrow$ \\
    \hline
    \hline 
     DA & 1.5969&   0.4039&   0.2444& \textbf{0.1593}&  \textbf{0.3791}&  0.6628 \\
    Ours & \textbf{1.3811}& \textbf{0.3969}&  \textbf{0.2304}&  0.0938&  0.3660&  \textbf{0.7482} \\
    \hline
    \hline
     DA &  4.3064&   0.5757&   0.4083&  0.0676&  0.1421&  0.2746 \\
     Ours & \textbf{3.8157}&  \textbf{0.5361}&  \textbf{0.3490}&   \textbf{0.0938}&  \textbf{0.1922}&  \textbf{0.3069} \\
    \hline
    \end{tabular}%
\end{center}
\end{table}%

\subsection{Ablation Studies}
\paragraph{Ablations about the Framework}

To further demonstrate the effectiveness of our framework, we use it to train other advanced UIE models. As shown in Table~\ref{tab:abla_framework}, we train UIEC$^2$Net and Ushape on the LSUI dataset, and NUICNet and IACC on the UIID dataset. To ensure a fair comparison, we keep all the model and training hyperparameters unchanged when integrating the model into our framework. We find that integrating these models into our framework yields significantly superior results compared to training them alone. For instance, PSNR increases by up to 2.4481 and SSIM increases by up to 0.0772. These experimental results indicate that our framework can be integrated with any advanced UIE model, consistently achieving performance enhancements.

\begin{table}[t]
\begin{center}
  \caption{Comparison of several advanced UIE models before and after integration into our framework. Red numbers indicate the performance improvements achieved by our framework.}
  \label{tab:abla_framework}%
    \begin{tabular}{c||c||ll}
    \hline
    Dataset & Method & \multicolumn{1}{c}{PSNR$\uparrow$}  & \multicolumn{1}{c}{SSIM$\uparrow$}  \\
    \hline
    \hline
    {\multirow{6}{*}{LSUI}} & UIEC$^2$Net & 25.0757 & 0.8708 \\
          & +Ours & 27.0192 (\textcolor{red}{+1.9435})& 0.9002 (\textcolor{red}{+0.0294})\\
\cline{2-4}          & UShape & {25.7630} & {0.8296}\\
          & +Ours & 27.8548 ({\textcolor{red}{+2.0918}}) & 0.9068 ({\textcolor{red}{+0.0772}}) \\
\cline{2-4}          & UIEConv  & 28.9155	& 0.9187 \\
          & +Ours & 29.9253 ({\textcolor{red}{+1.0098}}) & 0.9248 ({\textcolor{red}{+0.0061}}) \\
    \hline
    \hline
    {\multirow{6}{*}{UIID}} & NUICNet & {23.5712} & {0.8159} \\
          & +Ours & 25.2495 ({\textcolor{red}{+1.6783}}) & 0.8688 ({\textcolor{red}{+0.0529}}) \\
\cline{2-4}          & IACC  & 25.0095 & 0.8895\\
          & +Ours & 27.4576 (\textcolor{red}{+2.4481}) & 0.9167 (\textcolor{red}{+0.0272}) \\
\cline{2-4}          & UIEConv  & 27.4934 & 0.9129 \\
          & +Ours & 28.6446 (\textcolor{red}{+1.1512}) & 0.9203 (\textcolor{red}{+0.0074})\\
    \hline
    \end{tabular}%
\end{center}
\end{table}%
\begin{table}[t]
\begin{center}
  \caption{Ablation study of the UIEConv model structure.}
  \label{tab:abla_branch}%
    \begin{tabular}{c||cc||cc}
    \hline
     & \multicolumn{2}{c||}{LSUI}  & \multicolumn{2}{c}{UIEB} \\
    \hline
    Method & PSNR$\uparrow$ & SSIM$\uparrow$ & PSNR$\uparrow$ & SSIM$\uparrow$ \\
    \hline
    \hline
    Local Branch & 27.9657 & 0.9080 & 23.7615 & 0.9044\\
    Global Branch & 25.9986 & 0.8777 & 23.8900 & 0.8977 \\
    UIEConv & \textbf{28.9155} & \textbf{0.9187} & \textbf{24.1197} & \textbf{0.9283} \\
    \hline
    \end{tabular}%
\end{center}
\end{table}%

\begin{table}[t]
\begin{center}
  \caption{Ablation study of the depth degradation model (DDM) structure and loss function on the LUSI dataset.}
  \label{tab:abla_ddm}%
    \begin{tabular}{c|c||cc}
    \hline
    \multicolumn{2}{c||}{Method} & PSNR$\uparrow$ & SSIM$\uparrow$ \\
    \hline
    \hline
    \multicolumn{2}{c||}{Full Framework} & \textbf{29.9253} & \textbf{0.9248} \\
    \hline
    \multirow{2}{*}{VLEN}
    & w/o lowpass & 29.1634 & 0.9212 \\
    \cline{2-4}& w/o transform & 28.9733 & 0.9208 \\
    \hline 
    \multirow{3}{*}{DEN} 
    & finetune all & 29.5960 & 0.9221 \\
    \cline{2-4}& freeze all & 29.4437 & 0.9220 \\
    \cline{2-4}& scratch & nan   & nan \\
    \hline
    \multirow{3}{*}{FEN}
    & $\beta^D=\beta^B$ & 29.0706 & 0.9213 \\
    \cline{2-4}& w/o $z^{Scale},z^{Shift}$ & 29.1588 & 0.9211\\
    \cline{2-4}& w/o additional inputs & 29.4085 & 0.9218\\
    \hline
    \multirow{2}{*}{Loss} 
    & w/o $\mathcal{L}_{phy}$ & 28.7452 & 0.9191 \\
    \cline{2-4}& w/o $\mathcal{L}_{depth}$ & 29.6402 & 0.9227 \\
    \hline
    \end{tabular}%
\end{center}
\end{table}%

\paragraph{Ablations about UIEConv}
We perform an ablation study on the structure of UIEConv, as presented in Table~\ref{tab:abla_branch}. The performance significantly declines when either the global branch or the local branch of UIEConv is removed. This highlights the importance of both global features and local features for image enhancement. Consequently, the dual-branch structure of UIEConv is key to its superiority over other advanced UIE models.

\paragraph{Ablations about DDM and Loss Function}
We conduct a comprehensive ablation study on the design details of each sub-network within the DDM, as presented in Table~\ref{tab:abla_ddm}. "Full Framework" represents the complete version of our proposed framework. We remove each unique design element in the three sub-networks one by one to observe performance changes. For VLEN, we remove its low-pass filter (w/o lowpass) or the subsequent convolutional transformation (w/o transform). For DEN, we explore three different settings during training: fine-tuning its encoder and decoder (finetune all), freezing its encoder and decoder (freeze all), and training it from scratch without initializing with Depth Anything weights (scratch). For FEN, we experiment with several different structures. First, we retain one head to simultaneously estimate the attenuation coefficient $\beta^{D}$ and the scattering coefficient $\beta^{B}$ ($\beta^{D}=\beta^{B}$). Second, we remove the heads that estimate the scale $z^{Scale}$ and shift $z^{Shift}$ of the relative depth (w/o $z^{Scale}, z^{Shift}$), and instead use a predefined depth range (0.1m-10m) to scale the relative depth. Finally, we exclude the estimated veiling light $B^{\infty}$ and inverse depth $z$ as inputs to FEN (w/o additional inputs), using only the underwater image as the input. 
We find that removing any module or changing the training settings in the sub-networks leads to less accurate estimation performance, thereby reducing enhancement performance. Notably, training DEN from scratch causes the training process to fail to converge (resulting in NaNs), which indicates that the knowledge transferred from Pre-trained Depth Anything is crucial for the training of DEN and the entire framework.
Additionally, we study the role of two additional loss functions. Training without the physical constraint loss $\mathcal L_{phy}$ (w/o $\mathcal L_{phy}$) significantly decreased performance, highlighting the importance of physical guidance in our framework. Training without the depth consistency loss $\mathcal L_{depth}$ (w/o $\mathcal L_{depth}$) also slightly reduced performance.

\begin{figure*}[t]
\centering
\includegraphics[width=6.1in]{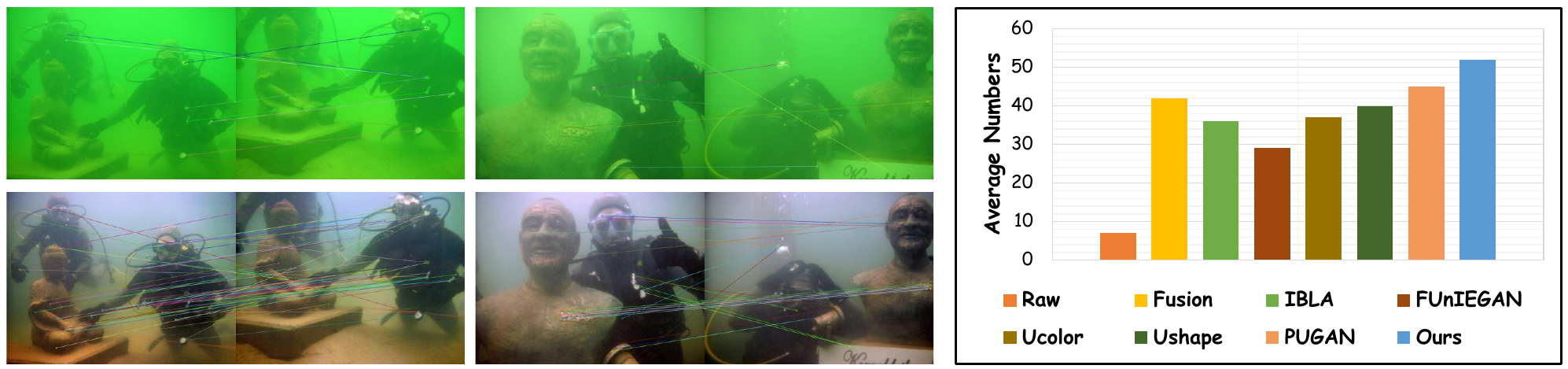}
\caption{The left panel shows keypoint matching results on raw underwater images and our enhanced images, while the right panel presents the number of matched keypoints for images enhanced using different methods.}
\label{fig:keypoint}
\end{figure*}

\begin{figure*}[ht]
\centering
\includegraphics[width=6.1in]{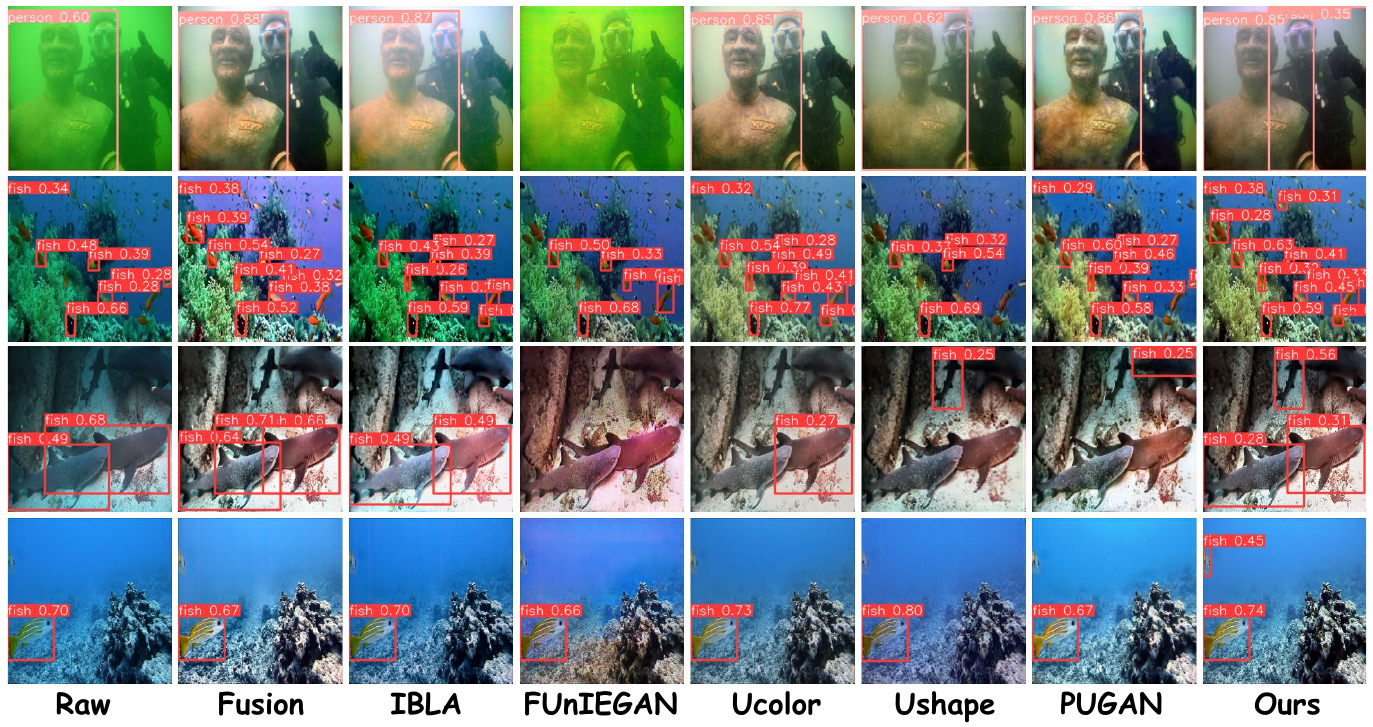}
\caption{Results of object detection on raw underwater images and images enhanced using different methods.}
\label{fig:detection}
\end{figure*}

\subsection{Application Tests}
To demonstrate the practicality of our method, we conduct two application tests. We use various UIE methods to enhance underwater images and perform keypoint matching and object detection on both the original and enhanced images. In Fig.~\ref{fig:keypoint}, we visualize the matched keypoints based on SIFT features. Compared to the raw images and the enhanced images from other methods, our enhanced images obtain the most keypoints. Additionally, we apply the YOLO-World model~\cite{YOLOWorld} for underwater object detection, setting the text prompts as \textit{person} and \textit{fish} to enable the detection of these two classes. Our method has the fewest missed detections in Fig.~\ref{fig:detection}.

\section{Limitations and Advantages}
The primary limitation of our method lies in its computational efficiency. As detailed in Tables~\ref{tab:sota} and \ref{tab:sota_deepsea}, our model's speed does not match some competing approaches (e.g., UGAN and FUnIEGAN), potentially restricting its direct application in real-time scenarios. However, our key advantage is a physics-model-guided training framework that ensures strong generalization across diverse underwater environments, including challenging non-uniform illumination (e.g., deep-sea conditions), where our method consistently achieves state-of-the-art performance. Furthermore, a significant byproduct of our image enhancement process is its ability to provide more accurate underwater scene depth estimations than specialized depth estimation algorithms.

\section{Conclusion}

In this paper, we introduce a novel physical model-guided framework for underwater image enhancement and depth estimation. This framework leverages the UIEConv model to enhance raw underwater images and employs the Deep Degradation Model (DDM) to estimate various parameters of the physical imaging model, including scene depth. By accurately simulating the degradation process, the physical imaging model bridges the relationship between the enhanced and raw underwater images, guiding the training of both UIEConv and DDM. Extensive experiments across diverse underwater environments validate the effectiveness of our framework, demonstrating significant improvements in both image quality and depth estimation. This robust framework offers a comprehensive solution to underwater imaging challenges, paving the way for further advancements in the field.



\bibliographystyle{IEEEtran}
\bibliography{bare_jrnl_new_sample4}

\end{document}